\documentclass[letterpaper]{article}
\usepackage{iccc}

\usepackage{epigraph}
\usepackage{graphicx}
\usepackage{subfig}  
\usepackage{booktabs}
\usepackage{makecell}
\usepackage{hyperref}
\usepackage{times}
\usepackage{helvet}
\usepackage{courier}
\usepackage{enumitem}
\usepackage{csquotes}
\title{Back to Back with a Copy: A Computational Analysis of \\
AI-Generated Visual Contemporary Art\textit{ Pastiches}}

\author{
Anca Dinu, Andreiana Mihail, Andra-Maria Florescu, Claudiu Creangă, Liviu Dinu\\
University of Bucharest, Bucharest, Romania\\
\texttt{anca.dinu@lls.unibuc.ro, andreianna@yahoo.com, andra-maria.florescu@s.unibuc.ro,}\\
\texttt{claudiudancreanga@gmail.com, liviu.p.dinu@gmail.com}
}

\begin{document} 
\maketitle
\begin{abstract}

The aim of this paper is twofold. First, it investigates whether newer generative models are getting \enquote{better} at \textit{pastiching }contemporary artworks. Second, it explores the consistency of the multidimensional nature of stylistic evaluation across different LLMs. 
Building on previous work, we analyze stylistic similarity between AI generated \textit{pastiches} and the original artworks of twelve contemporary artists. We used five complementary computer vision models to capture texture, color, semantics, composition, and perceptual features through cosine distance in high-dimensional embedding spaces. The distances obtained show that the newer image generation model that we used has produced \textit{pastiches} with improved semantic alignment and greater diversity than the model used in previous work. However, it was slightly less performant on shallow features such as color, texture, and perceptual adherence. Our findings confirm that artistic style is inherently multidimensional, and measuring it does not depend on any spatial architecture. 
These quantitative findings are contextualized through feedback from human evaluators, which are the artists themselves.

\end{abstract}


\section{Introduction}

Over the past decade, Computational Creativity (CC) has experienced rapid growth, largely driven by advances in AI, in particular transformer models, which greatly increased the capacity of computer systems to create original artifacts.

In this work, we focus on a type of visual computer creativity inherently combinatorial in nature, defined in \cite{boden2004creative} as the process of combining ideas from a given conceptual space in novel ways, which minds both convergent creativity (the capacity to identify the best or correct solution to a problem) and divergent creativity (the capacity to generate a wide range of varied original responses) \cite{guilford1950}, in an assisted creativity framework, defined in \cite{colton2012computational} as the automatic process of supporting a human creator. In particular, we asked the Gemini Pro 3 Nano Banana model \cite{geminiteam2025gemini} to  generate \textit{pastiches} after contemporary art, via prompting, by uploading the images of the original artworks. 

The starting point was the work of \cite{Dinu2026Art}, who employed Chat GPT to create \textit{pastiches} after the artworks of 12 contemporary artists. The study uncovered that the multi-faceted concept of style can be expressed by features like texture and color, semantics and concept, artistic touches, visual composition, or perceptions, and that these features can be captured independently using different computer vision models which extract high-dimensional embeddings. A large cosine distance between the AI generated \textit{pastiche} and the original human artwork in these embedding spaces means that the \textit{pastiche} was original, but satisfies less the task requirement of creating a \textit{pastiche}, while a low distance indicates a copy that is less original.

We kept the same framework and data, but we used the newer SOTA image generation  model Gemini. We were motivated by the following research questions: 
\begin{enumerate}
    \item Are newer generative models getting \enquote{better} at producing \textit{pastiches} after human artworks? 
    \item How do human and automatic evaluation correlate in this respect?
    \item Are visual models consistent in extracting stylistic features across different generative AI models?
    
\end{enumerate}

Our study directly compares the outputs of the two models and gains insight into the stability of the claims made in \cite{Dinu2026Art}. A vital part of our assessment was the contribution of the artists themselves through their evaluation and comparison.

\section{Related Work}

Recently, the fast growing body of work in CC spread to include a wide variety of topics, from improvement techniques and evaluation to ethical concerns and the role of humans in automatic image generation.

For example, \cite{morain2025prompt} showed that the use of more advanced prompt engineering techniques such as  Optimization as PROmpting (OPRO) or chain-of-thought (COT) leads to no significant improvements in textual computational creative systems, while \cite{colloca2025prompting} analyzed how ChatGPT amplifies positive sentiment and adds descriptive language regardless of the user's prompt style, showing no correlation between user prompt and AI-generated creative output.

The creative abilities of three artificial image-to-image generative models are formally quantified in  \cite{Ramaswamy2025CreativeBehavior}. The authors propose three objective creativity criteria: requirement satisfaction, cohesion, and novelty and introduce formal measures for each of them. The complexity and subjectivity of the evaluation process of CC systems, which depends mainly on the expertise and perception of the evaluator, are recognized in \cite{BODEN1996267,boden2004creative,Wiggins}. Moreover, \cite{DemkeVentura2024,morain2025prompt} point out that the domain is extremely diverse, making objective human evaluation difficult, time consuming, and costly.

Finally, \cite{Kaila2024GardeningFI} documented how artists negotiate authorship and ownership when experimenting with artificially generated imagery, revealing how they re-frame artistic roles and authorship through generative tools.

\section{Data}
\label{data}

This study uses the dataset proposed in \cite{Dinu2026Art}, which consists of 108 images: 

\begin{itemize}
    \item 36 original artworks by twelve contemporary artists, who generously agreed to lend 3 of their artworks each, for research purposes. The artists are: Adi Matei, Ciprian Mureșan, Ion Grigorescu, Iulia Uță, Karine Fauchard, Lazar Lyutakov, Marius Tănăsescu, Mathias Poeschl, Oana Năstăsache, Philip Patkowitsch, Răzvan Botiș, and Tom Chamberlain;
    \item 72 AI \textit{pastiches} produced by AI (ChatGPT 5 model), 2 imitations/\textit{pastiches} for each original. 
\end{itemize}

\begin{table}[t]
\centering
\small
\caption{Average Cosine Distances: Gemini vs GPT. Lower values indicate higher similarity.}
\label{tab:gemini_vs_gpt}
\renewcommand{\arraystretch}{1.2}
\begin{tabular}{l c c c l}
\hline
\textbf{Model} & \textbf{Gemini} & \textbf{GPT} & \textbf{$\Delta$\%} & \textbf{Feature Type} \\ \hline
AdaIN-Style & \textbf{0.064} & 0.063 & +1.6\% & texture, color \\
CLIP-ViT-L & \textbf{0.167} & 0.197 & -15.2\% & semantic \\
ResNet50-Style & \textbf{0.441} & 0.454 & -2.9\% & artistic style \\
DINOv2 & 0.461 & \textbf{0.463} & -0.4\% & f.g. visual \\
VGG19 & 0.699 & \textbf{0.674} & +3.7\% & perceptual \\ \hline
\end{tabular}
\end{table}

\begin{table}[t]
\centering
\small
\caption{Model Discrimination and Consistency: Gemini vs. GPT.}
\label{tab:variance_consistency}
\renewcommand{\arraystretch}{1.2}
\begin{tabular}{l c c c c}
\hline
\textbf{Model} & \multicolumn{2}{c}{\textbf{Variance}} & \multicolumn{2}{c}{\textbf{Consistency (Avg Diff)}} \\ \cline{2-5}
 & Gemini & GPT & Gemini & GPT \\ \hline
DINOv2 & 0.0425 & 0.0349 & 0.146 & 0.131 \\
VGG19 & 0.0082 & 0.0274 & 0.087 & 0.111 \\
ResNet50-Style & 0.0160 & 0.0249 & 0.089 & 0.115 \\
CLIP-ViT-L & 0.0072 & 0.0103 & 0.048 & 0.046 \\
AdaIN-Style & 0.0028 & 0.0014 & 0.029 & 0.034 \\ \hline
\end{tabular}
\end{table}

\begin{table}[t]
\centering
\caption{Extreme Divergence Cases: DINOv2 vs AdaIN. Divergence = DINOv2\textsubscript{norm} - AdaIN\textsubscript{norm}. Positive values indicate high texture adherence with high compositional divergence.}
\label{tab:divergence}
\renewcommand{\arraystretch}{1.2}
\begin{tabular}{l c c c}
\hline
\textbf{Artwork} & \textbf{DINOv2} & \textbf{AdaIN} & \textbf{Divergence} \\ \hline
Ciprian Mureșan 1 & 100\% & 20\% & +0.798 \\
Adi Matei 2 & 99\% & 27\% & +0.726 \\
Ion Grigorescu 3 & 81\% & 11\% & +0.697 \\ \hline
Tom Chamberlain 3 & 50\% & 97\% & -0.470 \\
Karine Fauchard 3 & 15\% & 42\% & -0.270 \\
Oana Năstăsache 2 & 5\% & 27\% & -0.213 \\ \hline
\end{tabular}
\end{table}

\begin{figure*}[htbp]
\section{Appendix}
    \centering
    \includegraphics[width=1.0\textwidth]{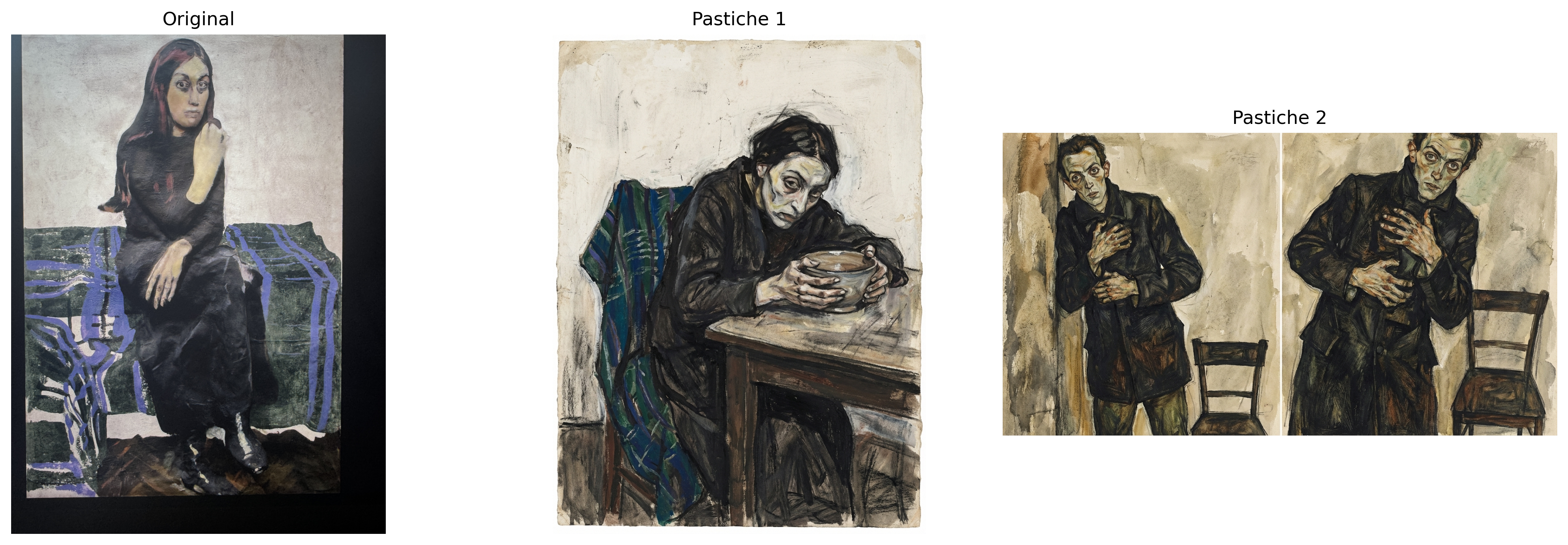}
    \caption{Examples of Gemini's \textit{pastiches} after Ion Grigorescu's work 1 \enquote{Măriuca Iosifescu}, oil on canvas and plaster, 1974.}
    \label{fig:ion_grigorescu_comparison}
\end{figure*}

\begin{figure*}[htbp]
    \centering
    \includegraphics[width=1.0\textwidth]{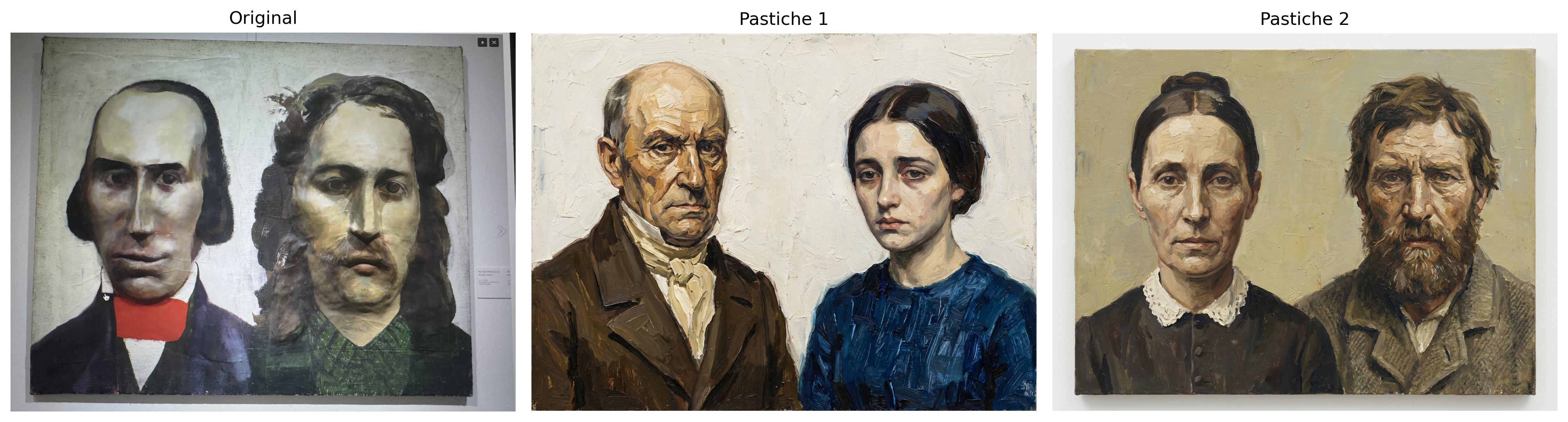}
    \caption{Examples of Gemini's \textit{pastiches} after Ion Grigorescu's work 2 \enquote{Avram Iancu și Nicolae Bălcescu}, oil on canvas and plaster, 1974.}
    \label{fig:ion_grigorescu_2_comparison}
\end{figure*}

\begin{figure*}[htbp]
    \centering
    \includegraphics[width=1.0\textwidth]{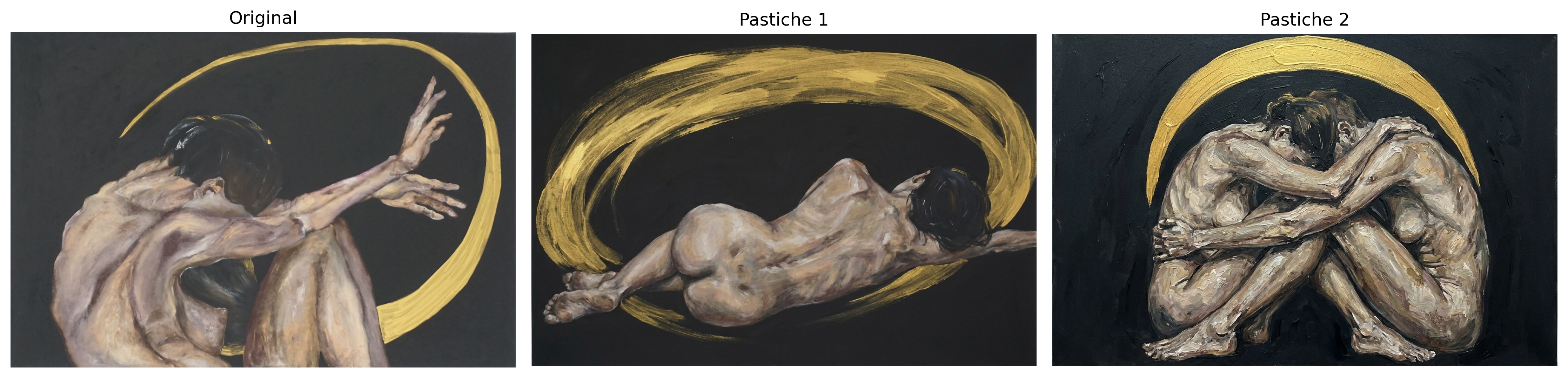}
    \caption{Examples of Gemini's \textit{pastiches} after Iulia Uța's work 2 Untitled, oil on canvas, 2024.}
    \label{fig:iulia_uta_2_comparison}
\end{figure*}

\begin{figure*}[htbp]
    \centering
    \includegraphics[width=1.0\textwidth]{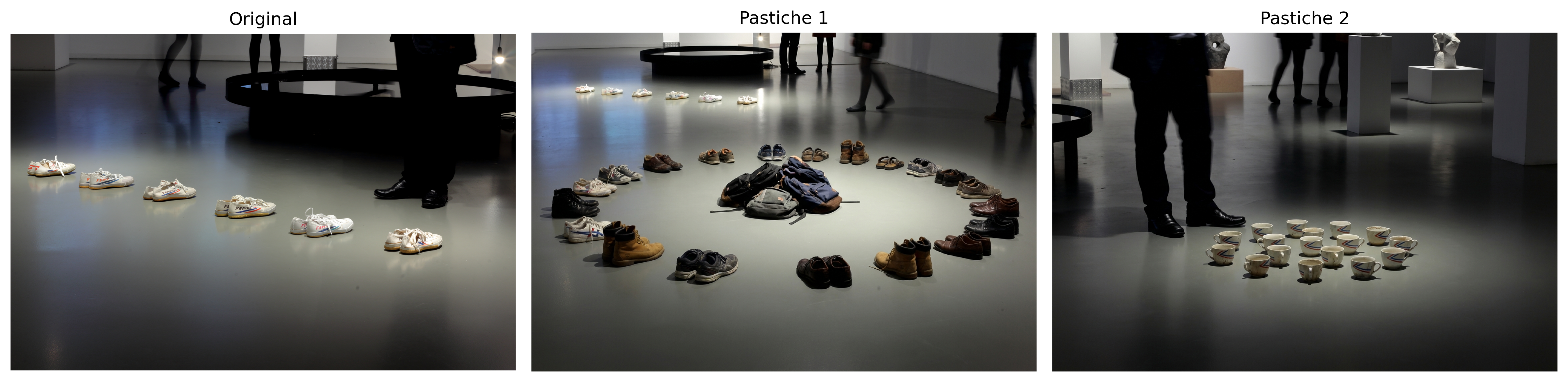}
    \caption{Examples of Gemini's \textit{pastiches} after Lazar Lyutakov's work \enquote{24/7 the human condition}, installation, 2014.}
    \label{fig:lazar_lyutakov_2_comparison}
\end{figure*}

We enriched this image dataset with a new series of AI generated works (e.g., Figures \ref{fig:ion_grigorescu_comparison}, \ref{fig:ion_grigorescu_2_comparison}, \ref{fig:iulia_uta_2_comparison}, \ref{fig:lazar_lyutakov_2_comparison} from the Appendix) using the latest Gemini Pro 3 Nano Banana model \cite{geminiteam2025gemini}. 
We used the same prompt as in the ChatGPT experiments conducted by \cite{Dinu2026Art}, so the results would be directly comparable.

\section{Experimental Setup}
\label{Experimentalsetup}

In the pre-reprocessing step, we converted all images to RGB format and normalized them according to each of the five models used in \cite{Dinu2026Art} to extract high-dimensional embeddings and which have been shown to capture different aspects of artistic style, as follows: 

\begin{itemize}
    \item AdaIN-Style \cite{Huang2017AdaIN} captures 1920-dimensions of texture/color statistics;
    \item ResNet50-Style  \cite{He2016ResNet} captures 2048-dimensions of artistic style features;
    \item CLIP-ViT-L \cite{Radford2021CLIP} captures 768-dimensions of semantic features;
    \item DINOv2 \cite{Oquab2023DINOv2} captures 1024-dimensions of fine-grained visual features;
    \item VGG19 \cite{Simonyan2015VGG} captures 4096-dimensions of perceptual features.
\end{itemize}

To approximate the stylistic similarity corresponding to each of the five models, for each artwork group (the original and its two Gemini-generated \textit{pastiches}), we compute three pairwise cosine distances: original to \textit{pastiche} 1, original to \textit{pastiche} 2, and\textit{ pastiche} 1 to \textit{pastiche} 2.

\section{Automatic evaluation}
\label{Results}

\begin{figure*}[htbp]
    \centering
    \caption{Average cosine distances between original artworks and Gemini-generated \textit{pastiches} across all five embedding models. Left panel shows Original$\rightarrow$\textit{Pastiche}~1 distances, middle shows Original$\rightarrow$\textit{Pastiche}~2, right shows \textit{Pastiche}~1$\leftrightarrow$\textit{Pastiche}~2 diversity. }
    \includegraphics[width=1.0\textwidth]{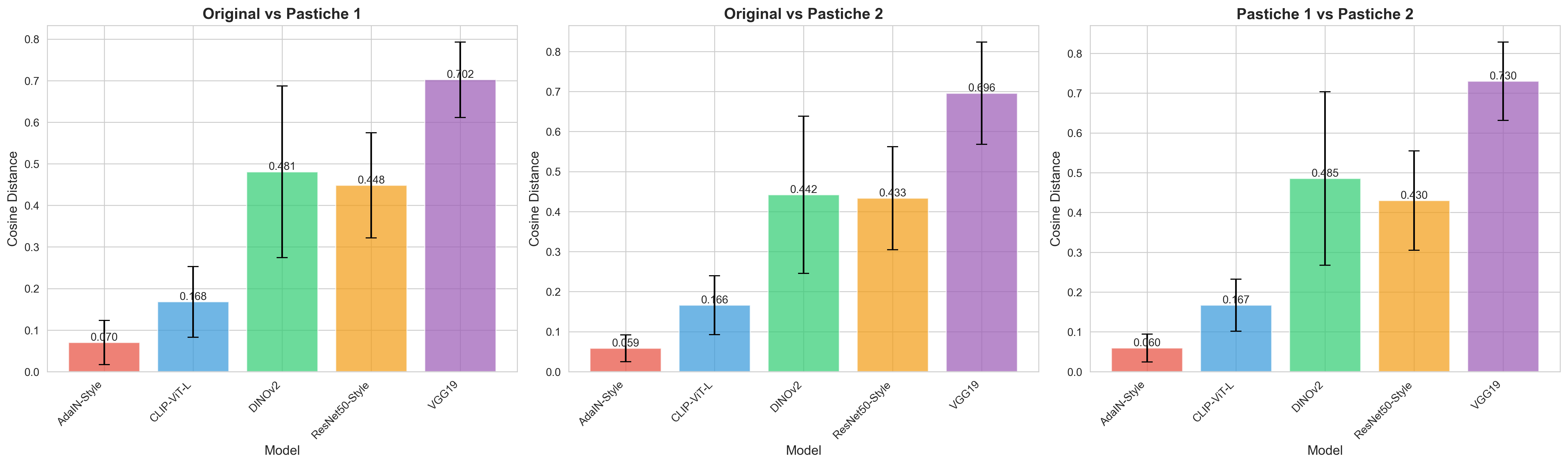}
    \label{fig:model_comparison}
\end{figure*}

The distribution of the cosine distance between the original artworks and their Gemini-generated \textit{pastiches} using the five distinct feature embedding models is given in Table \ref{tab:gemini_vs_gpt}, in comparison to the GPT results reported in \cite{Dinu2026Art}. Figure \ref{fig:model_comparison} illustrates these distances the Gemini generated \textit{pastiches}, for a more intuitive visual representation.

\begin{figure*}[htbp]
    \centering
        \includegraphics[width=1.0\textwidth]{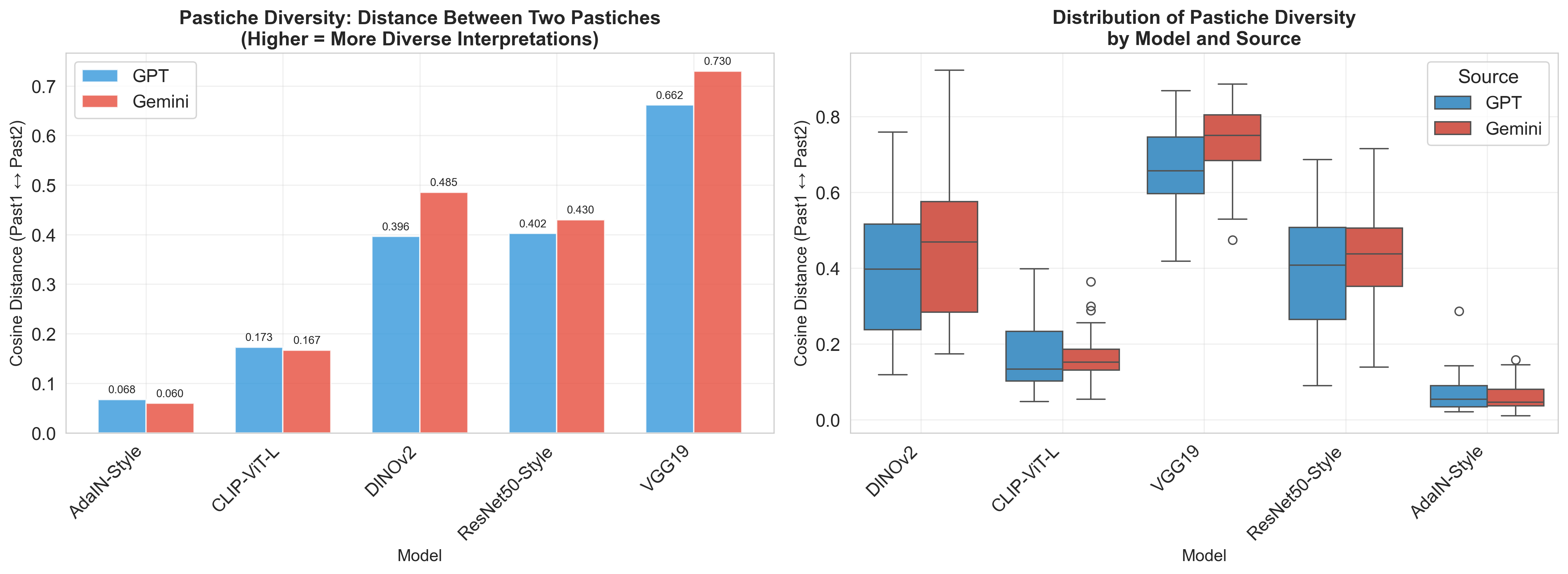}
        \caption{High \textit{pastiche} diversity (left) indicate the model explores more varied interpretations rather than producing near-identical outputs. The distribution boxplots (right)shows that Gemini produces overall 10.1\% more diverse \textit{pastiches}, with the largest differences in DINOv2 (+22.4\%) and VGG19 (+10.3\%), suggesting greater compositional and perceptual variety.}
    \label{fig:pastiche_consistency}
\end{figure*}

Comparing ChatGPT and Gemini reveals a consistent model distance hierarchy: AdaIN-Style remains the lowest (0.064), followed by CLIP-ViT-L (0.167), ResNet50-Style (0.441), DINOv2 (0.461), and VGG19 (0.699). This confirms that the \enquote{compositional gap} between texture matching and perceptual similarity is model-agnostic and inherent to the task itself. However, the models prioritize different elements (Table \ref{tab:gemini_vs_gpt}). Gemini favors deep semantics, better capturing concepts (15.2\% closer) and style (2.9\%), making it more adept at the overall \textit{pasticheing} task based on cosine distance. In contrast, ChatGPT focuses on surface forms, slightly better imitating perceptual features (3.7\% closer) and texture/color (1.6\%). Furthermore, Gemini generates significantly more diverse\textit{ pastiches}, as shown in Figure \ref{fig:pastiche_consistency} from the Appendix. 
The average distance between \textit{pastiches} of the same original is 10.1\% higher for Gemini than GPT (0.374 vs. 0.340). This diversity is particularly notable in DINOv2 (+22.4\%) and VGG19 (+10.3\%), indicating that Gemini explores broader compositional and perceptual interpretations.

We also compared the discrimination (variance) and consistency of the two models, given in Table \ref{tab:variance_consistency}. Discrimination reflects the embedding model's capacity to distinctly separate different artistic styles (measured by variance), while consistency indicates the stability of the generative model when producing multiple \textit{pastiches} for the same original artwork (measured by the average distance between them).

DINOv2 remains the most discriminative model for both Gemini (0.0425) and GPT (0.0349), while AdaIN-Style remains the least discriminative. Notably, Gemini shows improved consistency on VGG19 and ResNet50-Style compared to GPT, suggesting more stable perceptual and stylistic reproduction.

The models capture complementary information, as evidenced by the weak correlation ($r = 0.293$) between DINOv2 (composition) and AdaIN-Style (texture). This quantifies the \enquote{compositional gap}, showing that texture fidelity does not predict compositional accuracy. Table \ref{tab:divergence} highlights some extreme cases in which these metrics disagree. For example, Ciprian Mureșan's Work 1 shows the largest positive divergence (+0.798), pairing high texture adherence (AdaIN: 20th percentile) with high compositional divergence (DINOv2: 100th percentile). Conversely, Tom Chamberlain's Work 3 displays the largest negative divergence (-0.470), combining moderate compositional adherence (50th percentile) with high texture divergence (97th percentile).

\section{Artists' evaluation}

We asked artists to evaluate Gemini's \textit{pastiches} after their artwork, answering the same three questions about the \textit{pastiches} produced by ChatGPT in \cite{Dinu2026Art}, for a direct comparison:
\begin{enumerate}[label=Q\arabic*.]
    \item To what extent do you recognize your personal artistic language and the coherence of your visual style in this new work? (1 = not at all, 10 = completely)
    \item How does this new work inspire you or what thoughts does it provoke? (open answer)
    \item To what extent do you consider that the work generated by Gemini has aesthetic or artistic value? (1 = not at all, 10 = very high)
\end{enumerate}

\begin{figure}[htbp]
    \centering
    \includegraphics[width=0.45\textwidth]{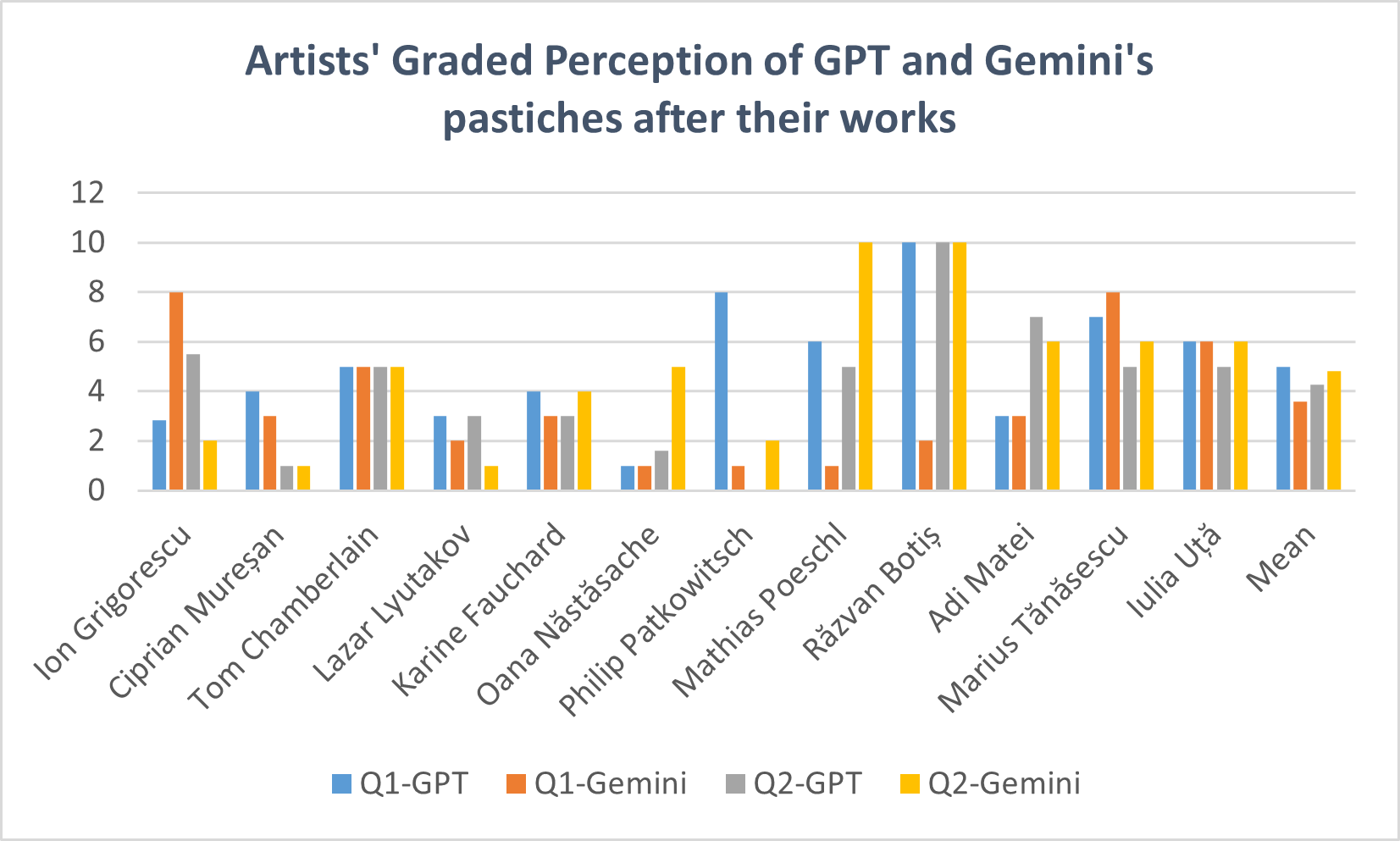}
    \caption{Comparison between artists' graded perception of GPT and Gemini's \textit{pastiches} after their works.}
    \label{fig:RaspunsuriArtisti}
\end{figure}

Similarly with ChatGPT's \textit{pastiches}, the grades for Gemini's \textit{pastiches} varied considerably by artist, as shown in Figure \ref{fig:RaspunsuriArtisti}. Compared to ChatGPT, Gemini's \textit{pastiches} were perceived by the artists as closer to their style, with a mean grade for the first question of 4.99, representing a significant increase of 1.41 points compared to the mean grade obtained by ChatGPT. However, they were perceived as slightly less artistically valuable, with a mean grade for the third question of 4.26, which is 0.67 points lower than the one obtained by ChatGPT's \textit{pastiches}. Overall, the artists seem to perceive Gemini's pastiches as closer to their works, in line with the computational evaluation, which constitutes a weak positive answer to our second RQ.

Answering the open-questions, the artists offered a variety of opinions about the \textit{pastiches} of their artworks. Most of them regarded the images created by AI more like \enquote{artistic comments} than real creations, with little understanding of originality, color mixing techniques, body, and depth.  Artistic comments, in this instance, refers to image that functions more as visual interpretation or reactions to the original artworks rather than independent creative works.

We reproduce here some of the answers. Oana Năstăsache says that \enquote{some colors look dirty(...) I would never associate that ochre-green combination, because it feels cold and muddy} or that \enquote{The surfaces are placed without any sense of composition}. Ion Grigorescu remarks that the new \textit{pastiches} followed the concept and composition, but failed in spirit, technique, and style: \enquote{Young man rising from a chair. Egon Schiele — closer to forgery than to \textit{pastiche}. A theatricality cheaper than Schiele’s}. Philip Patkowitsch has a different point of view raising the issue of  intellectual property: \enquote{they actually present my works better, but just because it is using parts of the original image, so I would say copyright infringement would be an issue!}.

So, in the end, did Gemini produce artworks as perceived by the artists themselves? Most of them responded negatively, motivating that AI displayed insufficient command of color theory, which resulted in a poor chromatic atmosphere and the lack of full conceptual comprehension/integration.

\section{Interpretation of the results}
\label{discussion}

\begin{figure*}[htbp]
    \centering
        \includegraphics[width=1.0\textwidth]{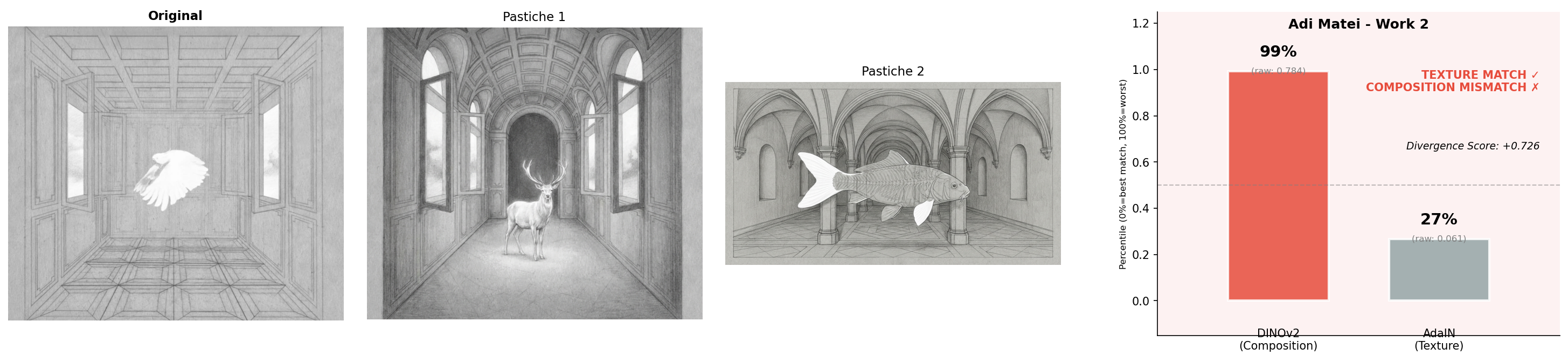}
    \caption{High texture adherence (low AdaIN), high compositional divergence (high DINOv2). These cases illustrate the core of the \enquote{compositional gap} and the orthogonality of style dimensions.}
    \label{fig:high_dino_low_adain}
\end{figure*}

\begin{figure*}[htbp]
    \centering
        \includegraphics[width=1.0\textwidth]{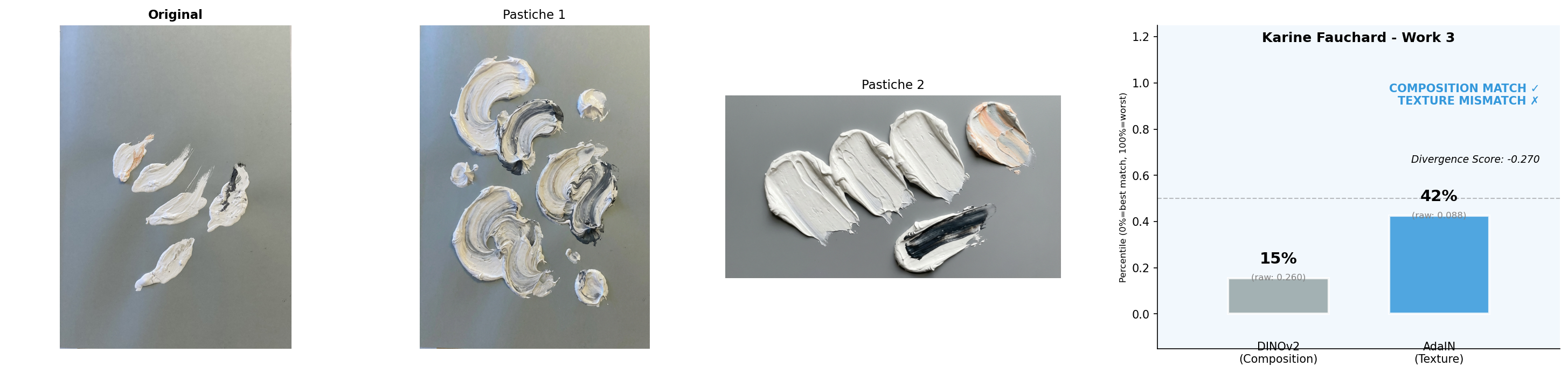}
    \caption{High compositional adherence (low DINOv2), high texture divergence (high AdaIN). Together with Figure~\ref{fig:high_dino_low_adain}, this confirms texture and composition as orthogonal dimensions.}
    \label{fig:low_dino_high_adain}
\end{figure*}

\begin{figure*}[htbp]
    \centering
    \includegraphics[width=1.0\textwidth]{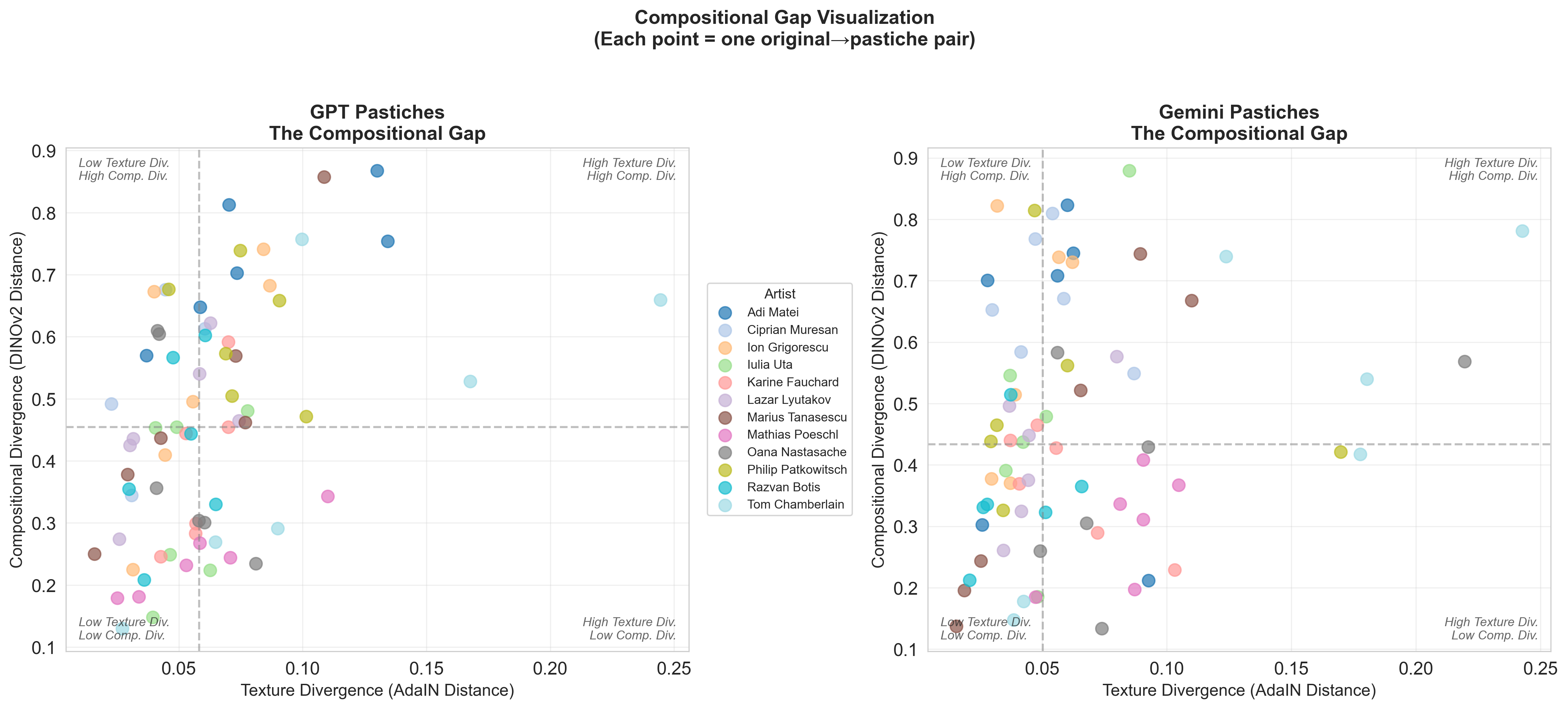}
    \caption{Texture divergence (AdaIN) vs. compositional divergence (DINOv2) for GPT (left) and Gemini (right) \textit{pastiches}. Each point represents one original→pastiche pair, colored by artist. Dashed lines indicate median values, dividing the space into four quadrants. Both models exhibit the same characteristic pattern: points cluster toward the left (low texture divergence, typically 0.03–0.10) but spread widely along the y-axis (compositional divergence ranging from 0.2 to 0.9). This asymmetric distribution—narrow horizontally, broad vertically—is the visual signature of the compositional gap: AI pastiches consistently adhere to texture/color statistics while showing high variability in compositional structure. The upper-left quadrant (\enquote{Low Texture Div. / High Comp. Div.}) is densely populated for both models, representing the dominant generation pattern. Artist-specific clustering is evident: Adi Matei (dark blue) and Ciprian Muresan (light blue) concentrate in high compositional divergence regions, while Oana Năstăsache (gray) and Mathias Poeschl (magenta) cluster toward lower compositional divergence. The similar distributions across GPT and Gemini confirm the compositional gap is model-agnostic.}   
    \label{fig:compositional_gap_scatter}
\end{figure*}

\begin{figure*}[t]
    
    \label{appendix:distributions}
    \vspace{1em}
    \centering
    \includegraphics[width=1.0\textwidth]{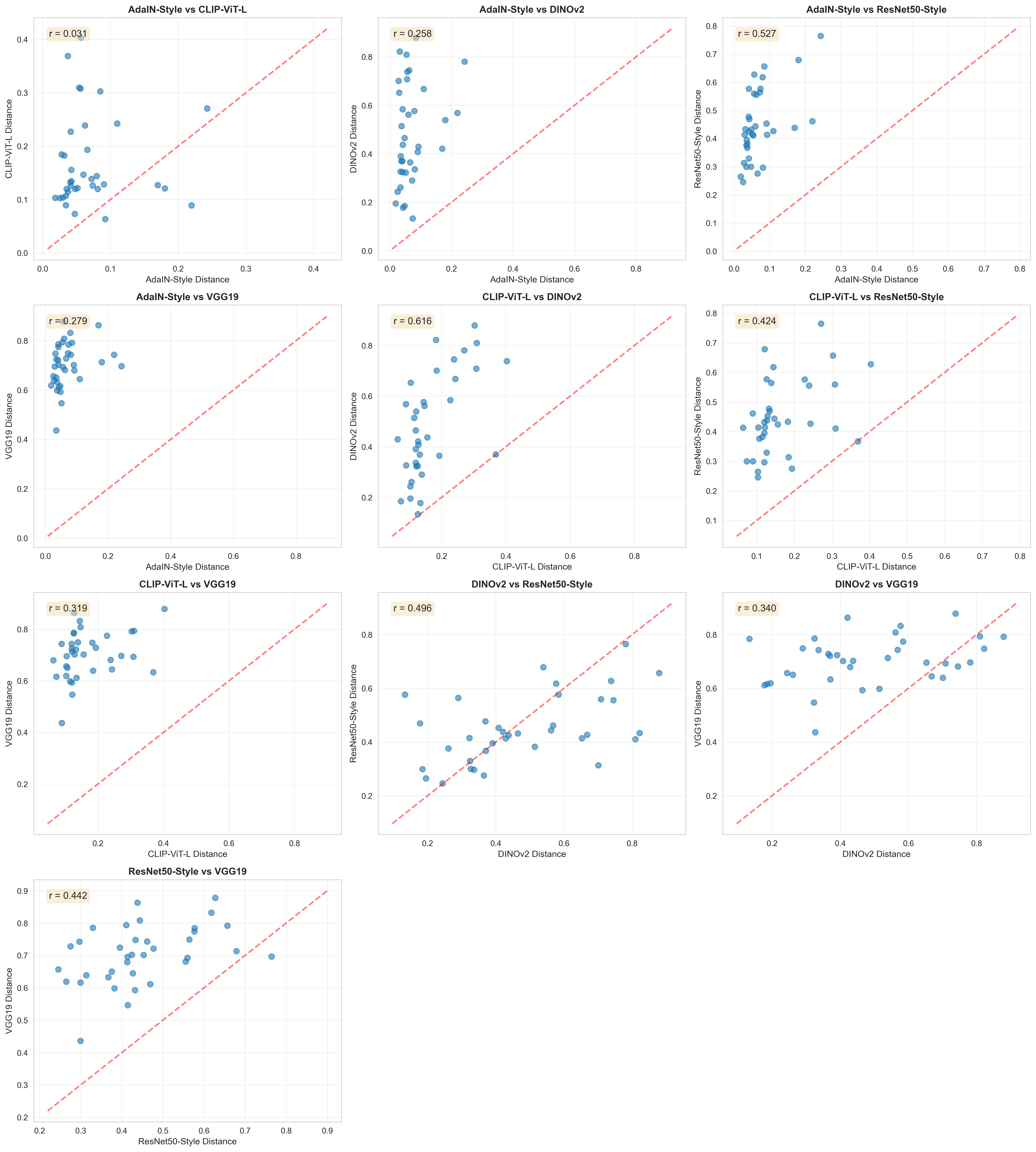}
    \caption{Pairwise correlations between embedding model distances. Each scatter plot shows the cosine distances from one model (x-axis) versus another (y-axis) across all 36 artist-\textit{pastiche} pairs, with the Pearson correlation coefficient ($r$) displayed. The red dashed line indicates the identity diagonal. Key findings: (1) AdaIN-Style and CLIP-ViT-L show near-zero correlation ($r = 0.031$), confirming that texture statistics and semantic content are orthogonal dimensions; (2) The highest correlation is between CLIP-ViT-L and DINOv2 ($r = 0.616$), reflecting their shared emphasis on higher-level visual features; (3) AdaIN-Style correlates most strongly with ResNet50-Style ($r = 0.527$), as both capture style-related properties; (4) VGG19 shows weak-to-moderate correlations with all other models ($r = 0.28$--$0.44$), suggesting it captures a distinct perceptual dimension. The generally weak correlations (mean $r \approx 0.37$) confirm that the five models capture complementary rather than redundant information, justifying the multi-metric evaluation framework.}
    \label{fig:model_corr}
\end{figure*}

The results confirm that artistic style should be evaluated multidimensionally \cite{Dinu2026Art}. Both GPT and Gemini successfully match texture statistics (low AdaIN distances) but show substantial compositional divergence (high DINOv2 and VGG19 distances), showing that these dimensions are orthogonal: high texture adherence often pairs with high compositional divergence, and vice versa, as illustrated in Figures \ref{fig:high_dino_low_adain}, \ref{fig:low_dino_high_adain}, and \ref{fig:model_corr} from the Appendix. Thus, this \enquote{compositional gap} is an intrinsic challenge for AI \textit{pastiche} generation, as visually shown in Figure \ref{fig:compositional_gap_scatter} from the Appendix, confirming that visual models consistently extract stylistic features across different generative AI models, positively respondig to our third RQ.

The models also exhibit distinct generation strategies, pointing to a weak positive answer to oure first RQ:

\begin{itemize}
    \item \textbf{Gemini}: An \enquote{interpretive} approach prioritizing semantic alignment (15\% closer to CLIP) and higher diversity (+10\%).
    \item \textbf{GPT}: A \enquote{faithful reproduction} approach prioritizing texture matching and consistency.
\end{itemize}

Because of the weak correlation ($r = 0.293$) between texture and composition, evaluating AI art requires a multi-metric framework:

\begin{itemize}
    \item \textbf{AdaIN-Style}: Texture/color fidelity.
    \item \textbf{CLIP-ViT-L}: Semantic/conceptual alignment.
    \item \textbf{DINOv2}: Compositional structure.
\end{itemize}

\section{Empirical Remarks}

We observed that Gemini often disregards prompt guidelines, defaulting to the styles of iconic masterpieces (e.g., Egon Schiele or Grant Wood's \textit{American Gothic}). Although this exploratory approach allows for originality, the model often overcompensates, producing works that compete with canonical art rather than complementing the original. Both models exhibit perfectionism, frequently completing works that artists intentionally left unfinished, while significantly altering the original image's foreground.

\begin{figure*}[htbp]

    \centering
     \includegraphics[width=0.20\textwidth]{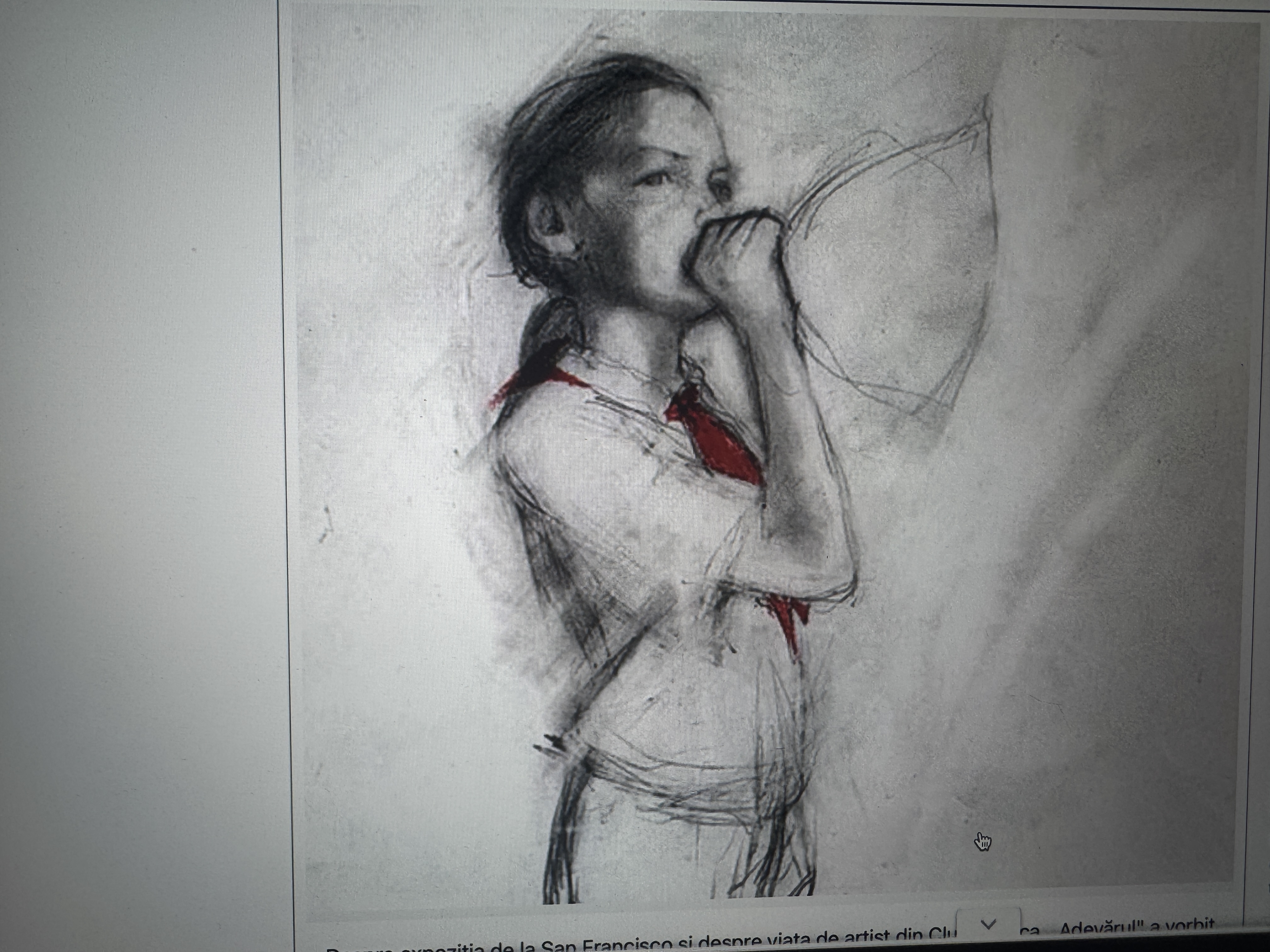}
     \hfill
    \includegraphics[width=0.20\textwidth]{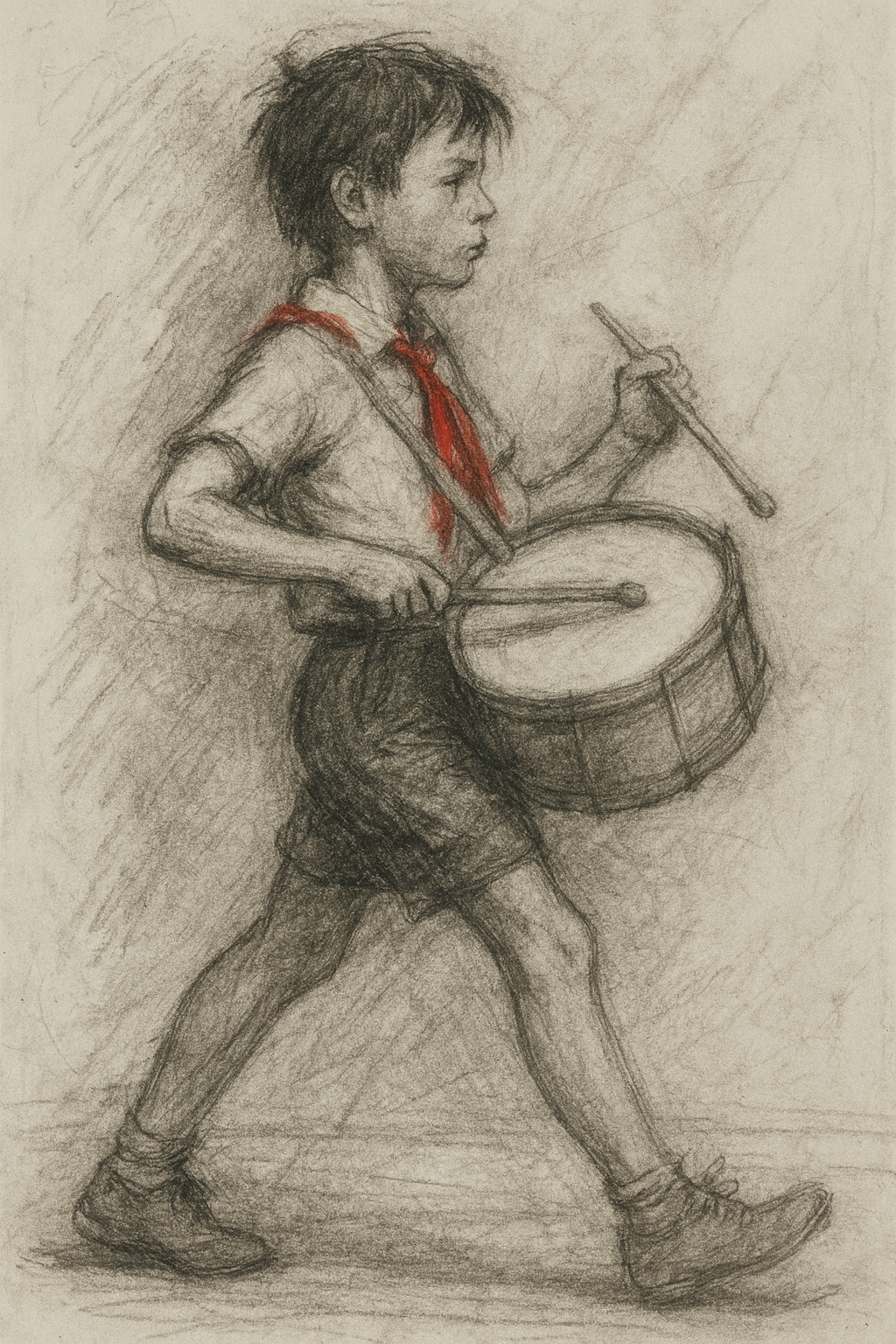}
    \hfill
    \includegraphics[width=0.30\textwidth]{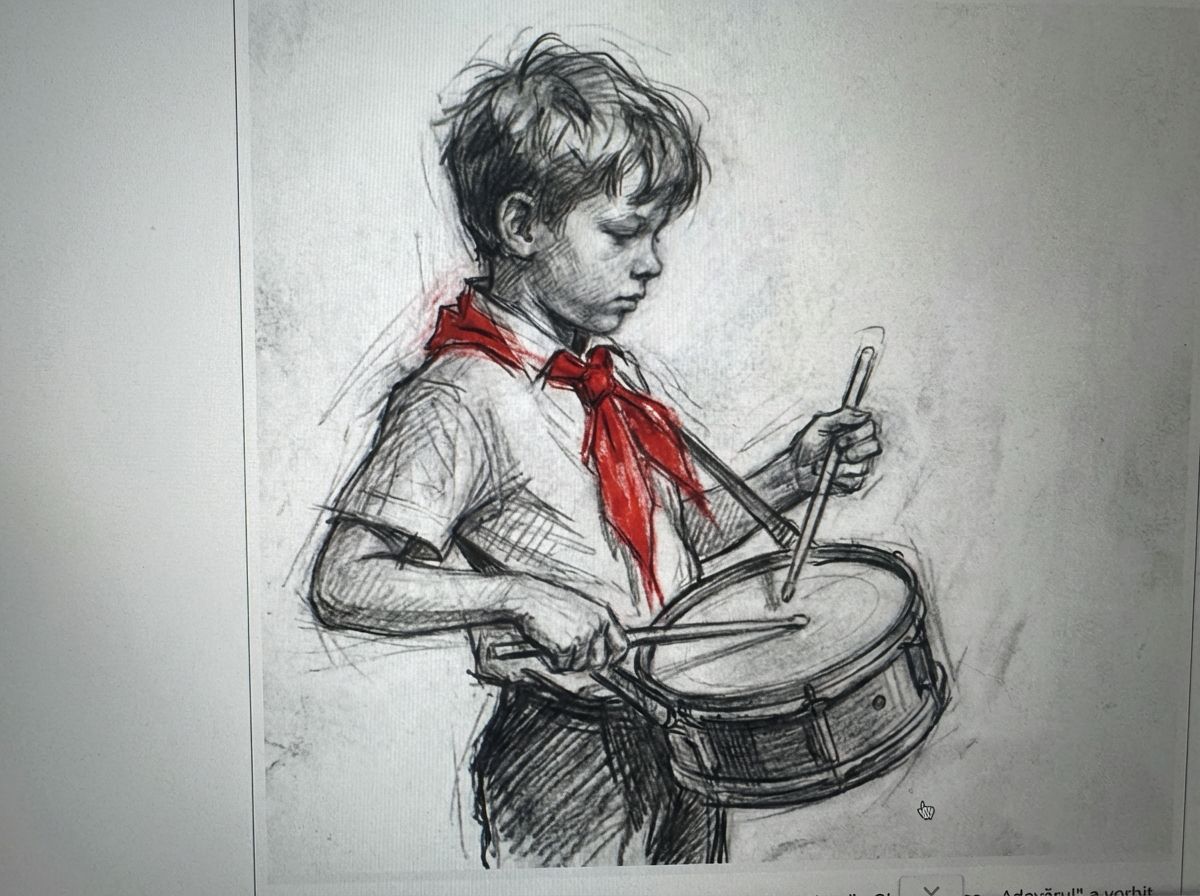}
    \caption{GPT (middle) and Gemini (right) created similar \textit{pastiches }of Mureșan's work (left)- suggest similar training data.}
    \label{fig:ciprian_muresan_comparison}
\end{figure*}

While Gemini \enquote{understands} dimensionality and brushwork, it consistently fails to perceive depth, context, and conceptual irony. A prime example is the \textit{pastiche} of Ciprian Mureșan's \textit{Pioneers}. Mureșan depicted communist youth as rebels and villains to mock 1980s Romanian propaganda. However, both Gemini and ChatGPT missed this irony, independently generating similar images of pioneers as standard patriotic drummers, shown in Figure \ref{fig:ciprian_muresan_comparison}. The striking similarity between the drummers generated by the two models suggests that they were trained on the same visual data. 


Finally, Gemini performs better when creating collages and installations compared to paintings. This combinatorial technique aligns with LLM capabilities, though the results often resemble over-accumulation rather than originality.

\section{Limitations and Future Work}
 
Our study is constrained by a small conceptually homogeneous dataset, as most artists belong to the same gallery. Including detailed figurative artwork could provide better insight into LLM \textit{pastiche} performance. Furthermore, comparing outputs across fundamentally different mediums (e.g., 2D paintings versus 3D conceptual installations) presents structural methodological challenges. Finally, photography was excluded because it functions as a mechanical reproduction of reality rather than an artwork.

We plan to adopt alternative metrics beyond the cosine distance that are better suited for evaluating visual similarity. To mitigate personal artist bias, we aim to incorporate formal evaluations by professional art critics. Lastly, we will apply this methodology to iconic, frequently \textit{pastiched} artworks, such as Da Vinci's \enquote{Mona Lisa,} Van Gogh's \enquote{Starry Night,} Grant Wood's \enquote{American Gothic,} and Andy Warhol's \enquote{Campbell's Soup Cans}.

\section{Ethical Implications}

This publication raises no ethical concerns, as all software licenses and artist permissions were strictly respected. However, broader generative AI (GenAI) art introduces significant socio-ecological challenges \cite{Marella_2025}. Societally, GenAI disrupts creative labor markets and raises issues regarding data exploitation, authorship, and intellectual property rights \cite{Frosio2023GenerativeAIinCourt}, often reinforcing existing inequalities\footnote{https://www.weforum.org/stories/2025/01/the-impact-of-genai-on-the-creative-industries/}. These disruptive impacts on cultural heritage and sustainability require a commitment to responsible inclusive cultural development \cite{Roe2025}.

\bibliographystyle{iccc}
\bibliography{iccc}

\textbf{Acknowledgments}

Research supported by a grant of the Ministry of Research, Innovation and Digitization, CNCS - UEFISCDI, project SIROLA, number PN-IV-P1-PCE-2023-1701, within PNCDI IV, and by UB project.

We are grateful to all the artists who agreed to let us use their works and for their insightful feedback: Adi Matei, Ciprian Mureșan, Ion Grigorescu, Iulia Uță, Karine Fauchard, Lazar Lyutakov, Marius Tănăsescu, Mathias Poeschl, Oana Năstăsache, Philip Patkowitsch, Răzvan Botiș, and Tom Chamberlain. 

\clearpage

\label{appendix:compositional-gap}

\label{appendix:embedding-space-visualisation}


\end{document}